# Using Statistical Models to Detect Occupancy in Buildings through Monitoring VOC, $CO_2$ and other Environmental Factors

**Mahsa Pahlavikhah Varnosfaderani,[1] Arsalan Heydarian, Ph.D [1], and Farrokh Jazizadeh, Ph.D.[2]**

[1]Link Lab, Department of Engineering Systems and Environment, University of Virginia, Charlottesville, VA, 22904; e-mail: mp3wp@virginia.edu, ah6rx@virginia.edu

[2]Department of Civil and Environmental Engineering, Virginia Polytechnic Institute and State University, Blacksburg, VA, 24061; e-mail: jazizade@vt.edu

**ABSTRACT**

Dynamic models of occupancy patterns have shown to be effective in optimizing building- systems operations. Previous research has relied on CO2 sensors and vision-based techniques to determine occupancy patterns. Vision-based techniques provide highly accurate information; however, they are very intrusive. Therefore, motion or CO2 sensors are more widely adopted worldwide. Volatile Organic Compounds (VOCs) are another pollutant originating from the occupants. However, a limited number of studies have evaluated the impact of occupants on VOC level. In this paper, continuous measurements of CO2, VOC, light, temperature, and humidity were recorded in a 17,000 sqft open office space for around four months. Using different statistical models (e.g., SVM, K-Nearest Neighbors, and Random Forest) we evaluated which combination of environmental factors provide more accurate insights on occupant presence. Our preliminary results indicate that VOC is a good indicator of occupancy detection in some cases. It is also concluded that a proper feature selection and developing appropriate global occupancy detection models can reduce the cost and energy of data collection without a significant impact on the accuracy.

## INTRODUCTION

Buildings are a major consumers of energy in the world, accounting for approximately 40% of the total energy consumption (D'Oca et al. 2018). Heating, ventilation, and air conditioning (HVAC) systems are considered as the highest contributors among building-related energy consumption sources and are responsible for approximately 48% of the total energy consumption in the buildings (Nasruddin et al. 2019). Consequently, HVAC systems can be one of the main targets to be considered when thinking of reducing energy consumption in buildings. Traditional HVAC systems operate based on the maximum design occupancy of a building during the occupied hours, which may cause an unnecessary increase in HVAC operations and ultimately increase in energy consumption (Wang et al. 2020). Meanwhile, reducing the ventilation rates have shown to result in occupants' discomfort (Yang et al. 2014), as well as increasing the chance for spread of harmful particles and viruses as we have learned more due to the COVID-19 pandemic.

In fact, traditional HVAC operations can lead to both energy waste and occupants' discomfort, which highlights the need for a change in their operations. To address this, having accurate information on building occupancy levels can be helpful in replacing the traditional HVAC operations with demand-response HVAC control (Feng et al. 2015; Yan et al. 2015). Demand-response HVAC control can significantly reduce the buildings' energy consumption during the occupied hours and off-hours by dynamically adjusting the air ventilation according to the occupancy-status and patterns of specific zones while preventing overcooling or over-heating of vacant zones (Jazizadeh et al. 2020; Jung and Jazizadeh 2019). Prior research has shown that by having access to accurate occupancy counts and patterns, building automation systems (BAS) can dynamically adjust and control the ventilation rates of the HVAC systems in different zones, resulting in up to 80% reduction in HVAC-related energy consumption (Brooks et al. 2015). Moreover, occupancy information is important for emergency evacuation (Filippoupolitis et al. 2016), security management (Chen et al. 2018), and controlling lighting systems (second highest source of energy consumption) of buildings (Zou et al. 2018a).

With the recent advancements in internet of things (IoT) along with ubiquitous computing, we can accurately develop occupancy model that can inform the operation of HVAC systems (Candanedo and Feldheim 2016; Mahsa and Mohammad 2017). Low cost and easy to deploy non-intrusive indoor environmental quality (IEQ) sensors such as $CO_2$, temperature, humidity, and light sensors have become widely available in modern buildings and shown to provide promising results on detecting occupancy counts and patterns (Zimmermann et al. 2018; Zou et al. 2018b) . Since occupants directly influence the pollutants in indoor environments by their presence, activities, and the products they use (e.g. perfume), using IEQ sensors can be a great help in detecting occupants' presence (Candanedo and Feldheim 2016).

Among the IEQ factors, $CO_2$ sensors are the most common modality that has been studied to identify occupancy estimation and detection. Although $CO_2$ has shown to be effective for occupancy detection, the slow spread of $CO_2$ in the environment and the time that it takes for $CO_2$ to build up will cause the results to have a time delay from the real building occupancy. Another limitation of $CO_2$ sensors is that there are other factors such as passive ventilation and sensor placement, that affect the $CO_2$ levels. As an example of the importance of $CO_2$ sensor placement, Pantelic et al. shows that occupants have a personal $CO_2$ cloud and if the $CO_2$ sensors are not adequately placed near the occupant's $CO_2$ cloud, their readings could be delayed or incorrect (Pantelic et al. 2020). As a result, the placement of these sensors is widely important, and many times not considered when such sensors are being placed in indoor environments.

In addition to $CO_2$, other environmental factors such as temperature, humidity, and light have also been used for occupancy detection tasks. Candanedo and Feldheim used environmental features including $CO_2$, light, temperature, and humidity and compared three learning algorithms, namely linear discriminant analysis (LDA), classification and regression trees (CART), and random forest (RF), in their occupancy detection system. They showed that satisfactory results can be obtained with using proper feature selection and learning methods (Candanedo and Feldheim 2016). In another work, Kraipeerapun and Amornsamankul used stacking for multiclass classification and binary detection of occupancy through utilizing $CO_2$, light, temperature, and humidity levels as input features (Kraipeerapun and Amornsamankur 2017).

Over the recent years, due to availability of more reliable commercially available IoT devices, monitoring other IEQ factors has also gained more attention. For instance, recent studies have shown that occupants' presence can have a significant impact on the changes in volatile organic compounds (VOCs). These changes could be as a result of presence of organic materials in indoor space such as food, cleanser or disinfectants, and aerosol sprays. However, to the best of the authors knowledge, no study has evaluated whether VOCs can provide similar or better occupancy-related information as other evaluated metrics such as $CO_2$, lighting, or temperature.

Besides sensor selection, training and testing approaches for the occupancy models can also influence the procedure and accuracy of predictive models. Occupancy models reported in previous research are typically tested in the same spaces that they were trained. However, this approach can face several limitations, such as being time-consuming and intrusive, as well as it is difficult or impossible to collect training data in some areas. To address this limitation, global occupancy models are recommended to be created, in which occupancy models are trained in one space and tested in another space (Yang et al. 2014).

To overcome the mentioned limitations in previous research, this paper evaluates whether VOC data is a reliable indicator of occupancy status of single- and multi-occupied spaces. Furthermore, we evaluate whether using different environmental factors including $CO_2$, VOC, light, temperature, and humidity can bring better insights on detecting occupancy presence. For this task, different statistical models (i.e., SVM, K-Nearest Neighbors, and Random Forest) are used and Random Forest feature selection technique substitutes the manual selecting of features which requires expertise and may miss some of the important features. This paper also provides global modeling accuracies for each of the spaces to see whether it is possible to use the training data of one of the spaces to detect occupancy in the other ones.

## METHODOLOGY AND DATA COLLECTION

The data collection for this study was conducted in Living Link Lab (L-LL) ("Living Link Lab" 2021) – a 17,000 sqft living lab located on the University of Virginia campus which includes approximately 250 occupants. L-LL consists of 24 single-occupancy offices, four conference rooms, an arena space, and three large open-layout spaces (Figure 1). This space is equipped with over 200 IEQ sensors, including temperature, humidity, light, motion, air quality ($CO_2$, VOC, PM2.5), and noise-level sensors. These sensors have been deployed over the past four years and as a result granular longitudinal data has been collected over this period.

For this study, we have chosen two single-occupancy offices (A and B) and one conference room, which included at least two motion sensors, light-level, temperature, humidity, and at least one air quality ($CO_2$ and VOC) sensors. The selected time period for data analysis were from January 15, 2020 to April 30, 2020 which includes both pre COVID-19, when space was at full occupancy, and post COVID-19 when the offices were not occupied, and HVAC systems were operating as normal. This helps us having enough data points for both occupied and non-occupied situations in the space. Since the space is considered a living lab, occupancy schedules were not pre-defined for single-occupied offices and the occupants in those rooms were living their normal lives. Ground truth for occupancy status was obtained using two motion sensors as well as occupants' calendar information, indicating when they were in and out of their office. Similarly, for the conference rooms, motion sensors along with the conference room's shared calendar were utilized to extract the ground truth for occupied versus non-occupied times.

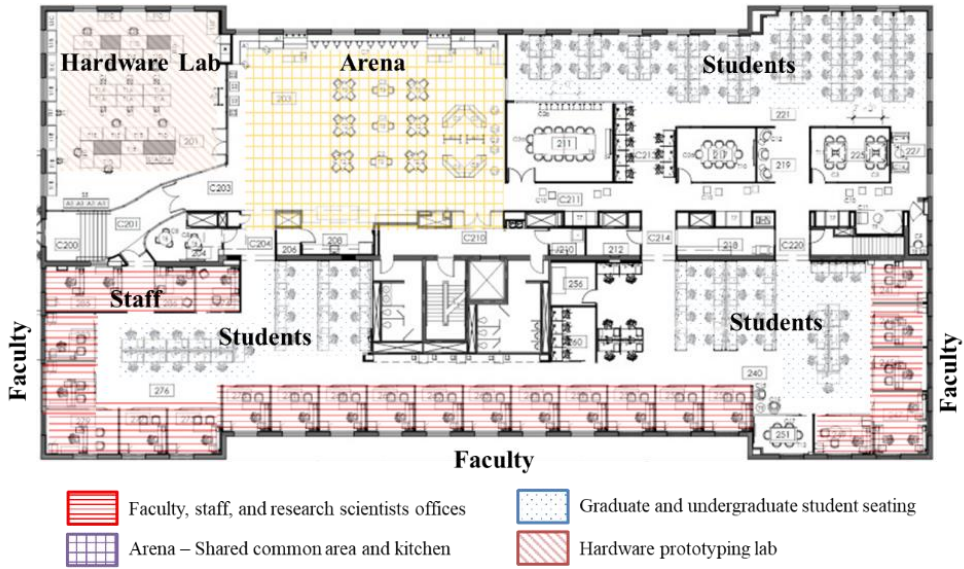

**Figure 1 - Living Link Lab space located at the University of Virginia**

In this paper, occupancy modeling is treated as a binary classification with possible outcome of 0 when there is no occupant and 1 when there is one or multiple occupants in the space. We adopted five machine learning models, including support vector machine (SVM), gaussian naive bayes (GNB), logistic regression (LGR), random forest classifier (RFC), and K-Nearest Neighbors (KNN) to identify which would fit the data better. Light, temperature, humidity, VOC, and $CO_2$ levels which were collected during the identified time frame were the input features for these algorithms. In the two single occupancy offices, two different $CO_2$ levels are considered, one with installing a sensor close to the occupants ($CO_2\_inhale$) and the other placed in the background area ($CO_2\_bg$) to evaluate the impact of $CO_2$ sensor placement.

For the SVM technique, C-Support Vector Classification with a kernel function is used. For training the model, we used both linear and non-linear kernels and chose the one that outperforms. Based on the results, when we are using more than two features for the occupancy detection task, using Gaussian kernel (as a non-linear kernel) leads to a higher accuracy. In this paper, radial basis function (RBF) is used as the Gaussian kernel. When training an SVM with the RBF kernel, there are two parameters that should be considered, the parameter C and gamma. Parameter C is common in all SVM kernels and trades of misclassification of training examples against simplicity of the decision surface. When C has a low value, it makes the decision surface smooth, while a high C aims at classifying all training examples correctly. For tuning the parameter C, we have used the cross-validation method and for parameter gamma in the RBF, that affects the decision boundary's flexibility, the following equation is used, in which *n* is the number of features.

$$\frac{1}{n \times X.var()}$$

Additionally, we tested the LGR model that can be used when the output of a problem is categorical, which in this case the occupancy detection is in a binary form. In this paper, the L-BFGS optimizer with L2 regularization is used for the LGR algorithm. The C parameter, which is the inverse of regularization strength is set to 10. This parameter is similar to the one in the SVM algorithm and the smaller values specify stronger regularization.

Neighbors-based classification was tested which is a type of instance-based learning that does not attempt to construct a general internal model and instead attempts to store instances of the training data. In this paper, we used the KNN model in which the classification is implemented by using the majority vote of the nearest neighbors for each data point. We used the uniform weights for this model. Parameter *k* in this model is the number of neighbors to use for queries and in general, larger values of *k* suppresses the effects of noise while making the classification boundaries less district. We have evaluated the model performance with *k* between 0 and 10 and set the parameter to 5 which outperforms other values. The distance metric parameter is set to Minkowski with power parameter of 2, that is equal to the Euclidean distance.

Furthermore, the Naive Bayes (NB) method was tested which is a set of supervised learning algorithms based on Bayes' theorem with the "naive" assumption of conditional independence between every pair of features given the value of the class variable. Bayes' theorem states the following relationship:

$$P(y \mid x_1, \ldots, x_n) = \frac{P(y)P(x_1, \ldots, x_n \mid y)}{P(x_1, \ldots, x_n)}$$

in which y is the category variable and $x_1$ through $x_n$ is the feature vector. Based on the strong assumption in NB about the independence of all features, we used the Maximum A Posteriori (MAP) estimation the Gaussian NB (GNB) was considered for this classification task.

Lastly, we tested the Random forests (RF) decision tree (DT), in which each of these DTs spits out a class prediction (in classification cases) and then the prediction of the RF model will be the class with most votes. RFs are generally more accurate than each of its DTs, by using the fundamental concept of wisdom of crowd. In this paper we used RF classifier (RFC) for the occupancy detection and feature selection task.

To test the accuracy of global models, we trained one set of models using the IEQ and occupancy data from one of the single-occupancy offices and used them to predict the occupancy of the two single occupancy spaces and the conference room. Another global model was trained using the data collected from the conference room and was tested to predict the occupancy of the two single-occupancy offices. We expect the global model created from the single occupancy office would perform better than the one created from the conference room data. The data was divided into training and testing sets for which we used the 10-fold cross-validation resampling method.

## RESULTS:

As shown in Figure 2, both VOC and $CO_2$ levels significantly change upon the presence of occupants. For evaluating each of the mentioned models, we have used individual environmental features and a combination of them to evaluate the impact of the features on the accuracy of each of the models. Table 1 shows the results of using the predictive models for occupancy detection in all three offices. As shown in the results across all predictive models, for single occupancy offices, using $CO_2$-bg (background $CO_2$) decreases the accuracy by around 10% compared to the $CO_2$_inhale. As we can see in the results of both single-occupancy offices, $CO_2$ is a good indicator of occupant's presence in both offices, but VOC is a good indicator only in office A. The reason for this can be that people emit $CO_2$ by breathing while their activities and the products they use/wear are sources of VOC. Therefore, the performance of VOC as an indicator of presence of the occupants in the space is greatly influenced by the occupants' characteristics and profile.

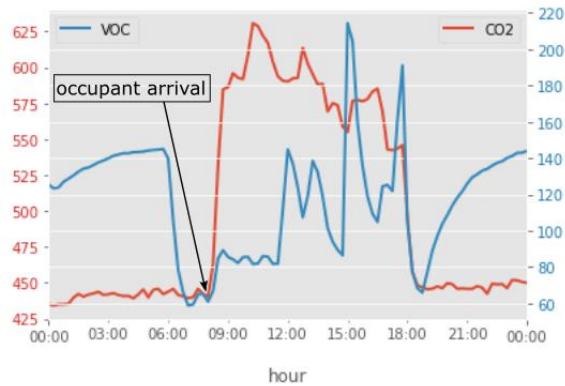

**Figure 2. VOC and $CO_2$ changes**

**Table 1. Local models' accuracies for different offices**

| Office A / Office B / Conference Room | | | | | |
|---|---|---|---|---|---|
| **Feature** | **SVM** | **GNB** | **LGR** | **RFC** | **KNN** |
| $CO_2$_inhale | 84 / 88 / - | 78 / 87 / - | 79 / 87 / - | 79 / 87 / - | 76 / 81 / - |
| $CO_2$_bg | 75 / 75 / 64 | 64 / 72 / 65 | 64 / 73 / 64 | 67 / 70 / 69 | 66 / 66 / 61 |
| VOC | 76 / 63 / 56 | 71 / 50 / 30 | 66 / 46 / 39 | 68 / 44 / 50 | 68 / 52 / 56 |
| Light | 76 / - / 82 | 56 / - / 80 | 61 / - / 81 | 70 / - / 79 | 66 / - / 78 |
| Temperature | 76 / 63 / 61 | 55 / 58 / 69 | 57 / 58 / 69 | 57 / 56 / 70 | 61 / 52 / 62 |
| Humidity | 76 / 63 / 64 | 45 / 45 / 61 | 42 / 47 / 60 | 50 / 48 / 51 | 61 / 50 / 60 |
| $CO_2$+VOC | 85 / 88 / 74 | 79 / 86/ 69 | 83 / 87 / 68 | 84 / 86 / 71 | 86 / 85 / 65 |
| $CO_2$+VOC+light | 85 / - / 82 | 80 / - / 81 | 83 / - / 81 | 84 / - / 80 | 86 / - / 70 |

In office A, VOC is a better indicator of occupancy than the background $CO_2$. The conference room does not get any sunlight and the best feature showing the occupancy status is light, since the lighting is controlled by the occupants. In the Link Lab, the control of occupants on the temperature is limited and they have no control on humidity, and as we can see the results show that these features are not good indicator for detecting occupancy presence in all three rooms, and they have been removed by RF feature selection method. In the conference room, we can see that $CO_2$ and VOC individually are not good indicators of occupancy; however, when combined, they improved the models' performance by up to 10%.

The results of the global model trained with the data from a single occupancy room showed great accuracy for detecting occupancy in the other single-occupancy office. It showed at most 2% decrease in the models' accuracies compared with the locally trained models. However, using the same global model to detect occupancy in the conference room indicates an up to 10% drop compared with the locally trained models from the same space.

**CONCLUSION AND FUTURE WORK**

In this paper, we used different statistical models to detect the occupancy status of two single-occupied and one multi-occupied offices using different environmental features. These modellings showed that VOC can sometimes work better than background $CO_2$ for detecting the occupancy level and can be used in combined with $CO_2$ to improve the occupancy predictions. The other finding is that global models can be good occupancy detectors in case the trained and test data are collected from offices with similar occupancy capacity. The results also showed that the location of $CO_2$ sensors is significantly important in detecting the occupants' presence. The performance of VOC in detecting occupancy can be dependent on the occupants' activities and the products they use, as these are the ways by which occupants emit VOC. A future direction of this research is to go one step further than occupancy detection and use the CO2 and VOC data for detecting occupants' activities as well as the different events that are happening in indoor environments. Another future vision of this work is to use the IEQ factors to estimate the number of occupants in the space.

**ACKNOWLEDGEMENT**

This research was supported in part by the National Science Foundation under grant #1823325. Special thanks to Alan Wang and Jacob Rantas, Ph.D. candidate and undergraduate researcher, respectively, who assisted with the installation the sensors, and data management. Our gratitude also goes out to all the residents of UVa Link Lab, for supporting the Living Link Lab research infrastructure.